\documentclass{article} % For LaTeX2e
\usepackage{iclr2025_conference,times}

\usepackage{amsmath,amsfonts,bm}

\def\eqref#1{equation~\ref{#1}}
\def\1{\bm{1}}

\def\va{{\bm{a}}}
\def\vb{{\bm{b}}}

\def\ve{{\bm{e}}}

\def\vp{{\bm{p}}}

\def\vr{{\bm{r}}}
\def\vs{{\bm{s}}}

\def\vx{{\bm{x}}}
\def\vy{{\bm{y}}}
\def\vz{{\bm{z}}}

\DeclareMathAlphabet{\mathsfit}{\encodingdefault}{\sfdefault}{m}{sl}
\SetMathAlphabet{\mathsfit}{bold}{\encodingdefault}{\sfdefault}{bx}{n}

\usepackage{hyperref}
\usepackage{url}
\usepackage{enumitem}

\usepackage{amsthm}
\usepackage{caption}
\usepackage{listings}
\usepackage{graphicx}
\graphicspath{ {./figures/} }

\usepackage{pgffor}
\usepackage{ifthen}

\title{Mechanism and Emergence of Stacked Attention Heads in Multi-Layer Transformers}

\author{Tiberiu Mușat \\
ETH Zürich, Switzerland \\
Giotto.ai, Switzerland\\
\texttt{tmusat@ethz.ch}
}

\iclrfinalcopy % Uncomment for camera-ready version, but NOT for submission.
\begin{document}

\maketitle

\begin{abstract}
   In this paper, I introduce the \textit{retrieval} problem, a simple yet common
   reasoning task that can be
   solved only by transformers with a minimum number of layers,
   which grows logarithmically with the input size.
   I empirically show that large language models can solve
   the task under different prompting formulations without any fine-tuning. To
   understand how transformers solve the retrieval problem, I train
   several transformers on a minimal formulation. Successful
   learning occurs only under the presence of an implicit curriculum. I uncover
   the learned mechanisms by studying the attention maps in the trained transformers.
   I also study the training process, uncovering that attention heads always
   emerge in a specific sequence guided by the implicit curriculum.
\end{abstract}

\begin{figure}[h]
   \begin{center}
      \includegraphics[width=5.4in]{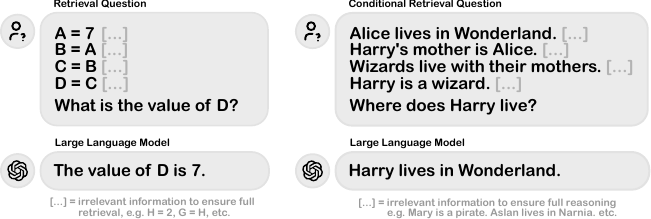}
   \end{center}
   \caption{Illustrative examples of \textit{retrieval} and \textit{conditional
         retrieval} questions.}
\end{figure}

\section{Introduction}

How do neural networks solve the tasks that they are trained on? Is there a
clear algorithm hiding behind the millions of unintelligible weights and
biases? These are the questions that the field of mechanistic interpretability
tries to answer. If successful, this line of research could lead to a better
understanding of neural networks and the development of AI systems with
increased safety, reliability, and efficiency \citep{doshi2017towards,
   olah2020zoom, elhage2021mathematical}.

Transformers \citep{vaswani2017attention} have become the dominant architecture
in natural language processing, achieving state-of-the-art results on a wide
range of tasks \citep{brown2020language, achiam2023gpt}. Recent
interpretability research has successfully uncovered the mechanisms learned by
single-layer \citep{nanda2023progress, quirke2023understanding} and two-layer
\citep{olsson2022context} transformers. However, understanding the mechanisms
learned by deeper transformers remains an open problem. Automatic circuit
analysis of large language models provides valuable insights about isolated
circuits that span several layers, but such circuits remain not fully
understood \citep{wang2022interpretability, conmy2023towards}. Therefore,
understanding the mechanisms of multi-layer transformers is a crucial step
towards understanding state-of-the-art language models.

\section{My Contribution}

\begin{samepage}
   In this paper, I try to answer the following questions:

   \begin{enumerate}[label=\textbf{Q\arabic*.}, itemsep=1pt]
      \item Are there tasks that can be solved only by transformers with a specific depth?
      \item Are large language models able to solve such tasks without specific
            fine-tuning?
      \item What is the mechanism that transformers use to solve the task?
      \item How does this mechanism emerge during training?
   \end{enumerate}
\end{samepage}

I answer \textbf{Q1} positively by introducing the \textit{retrieval} problem,
as well as a close variant that I term the \textit{conditional retrieval}
problem (Section \ref{sec:definition}). I answer \textbf{Q2} positively in
Section \ref{sec:llms} by empirically showing that large language models can
solve both problems without any specific fine-tuning under multiple prompting
formulations. In Section \ref{sec:theory}, I provide a formal proof that the
retrieval problem requires a minimum number of transformer layers that is
logarithmic in the input size.

This suggests that large language models have learned a complex mechanism
formed by multiple stacked attention heads. To elucidate this mechanism
(\textbf{Q3}), I train several transformers on a minimal formulation of the
retrieval problem (Sections \ref{sec:minimal-formulation} and
\ref{sec:implicit-curriculum}). In Section \ref{sec:reverse_engineering}, by
studying the attention maps in the trained transformers, I uncover multiple
possible mechanisms that I term \textit{retrieval heads}. Regarding the
training process (\textbf{Q4}), I find that retrieval heads emerge only under
the presence of an implicit curriculum and always in a specific sequence
(Section \ref{head_emergence}).

The retrieval problem also has important implications for \textbf{emergent
   abilities} in large language models \citep{wei2022emergent}, which I discuss in
Section \ref{sec:discussion}.

\section{The Retrieval Problem}
\label{sec:definition}

\subsection{Definition}

The \textit{retrieval} problem is directly inspired by the \textit{induction}
problem introduced by \citet{olsson2022context}. Given a sequence of tokens
$\ldots\va\vb{\ldots}\va$, the induction problem requires the model to predict
the token $\vb$. I directly extend this formulation by increasing the number of
induction steps to $D$. Given an input sequence
$\ldots\vx_{D-1}\vx_D\ldots\ldots{\ldots}\vx_1\vx_2{\ldots}\vx_0\vx_1{\ldots}\vx_0$,
the retrieval problem consists in predicting the token $\vx_D$. By setting
$D=1$, we recover exactly the original induction problem. Throughout this
paper, I also refer to the tokens in the retrieval chain using capital letters
(i.e., \textbf{A} for $\vx_0$, \textbf{B} for $\vx_1$, \textbf{C} for $\vx_2$,
and so on).

I also propose a more general variant of the retrieval problem, which I term
the \textit{conditional retrieval} problem, where each retrieval step could
depend on multiple previously retrieved values, not just the last one. For
example, given the input sequence
$\ldots\vx\vy\vz{\ldots}\va\vy{\ldots}\va\vx{\ldots}\va$, predicting the token
$\vz$ would constitute a conditional retrieval problem. The retrieval steps in
the retrieval problem are perfectly linear, while in the conditional retrieval,
they form a directed acyclic graph.

\subsection{Motivation}

The retrieval problem is implicitly present as a subproblem in many common
language tasks such as working with relations between persons, tracking the
evolution of a concept, solving mathematical and reasoning problems,
programming, and many more. Consider the following real-world example from the
Wikipedia article on llamas:

\textit{``\textbf{Llamas} are social animals and live with others as a
\textbf{herd}. {\textup{[\,\dots]}} A \textbf{cria} (from Spanish for `baby')
is the name for a baby \textbf{llama}, alpaca, vicuña, or guanaco.
\textbf{Crias} are typically born with all the females of the \textbf{herd}
gathering around.''}

An autoregressive language model trying to predict the second occurrence of the
word \textit{herd} (rather than \textit{flock}, \textit{group}, or
\textit{pack}) would need to first retrieve the fact that crias are llamas, and
then use it to retrieve the fact that llamas live in herds. This process is
essentially a retrieval problem with $D=2$.

From one point of view, the retrieval problem is essentially about working with
relations between entities, which is fundamental for language and reasoning.
This makes it an ideal testbed for studying the inner workings of large
language models.

\section{Large Language Models}
\label{sec:llms}

To better illustrate the task and to enable benchmarking of large language
models, I propose 5 specific formulations of the retrieval problems: 3
\textit{retrieval} formulations and 2 \textit{conditional retrieval}
formulations.

\begin{enumerate}[label=\textbf{F\arabic*.}, itemsep=1pt]
   \item \textit{Equations} formulation: ``a = 3. b = a. c = b. c = ?''
   \item \textit{Lives-with} formulation: ``Alice lives in Switzerland. Bob lives with Alice. Charlie lives with Bob. David lives with Charlie. Where does David live?''
   \item \textit{Kingdoms} formulation: ``Alice lives in Novaria. Novarians believe in harmonianism. Harmonianists eat lamb. Lamb contains Zephyrium. Zephyrium causes Chronogy. Who has Chronogy?''
   \item \textit{Functions} formulation (conditional retrieval): ``f(2) = 3. g = f. a = 2. g(a) = ?''
   \item \textit{Relatives} formulation (conditional retrieval): ``Jane lives in Switzerland. Alex's mother is Jane. Engineers live with their mothers. Alex is an engineer. Where does Alex live?''
\end{enumerate}

To ensure that the retrieval problem is not trivially solvable by just finding
the noun that fits the question, I interleave multiple retrieval chains in the
same question. This ensures that the model performs the entire reasoning chain.
To facilitate benchmarking, I also ask the model to output the answer directly
without any additional words and I repeat sampling until an acceptable answer
is generated. In Appendix \ref{prompts}, I provide examples of the full
prompts, correct answers, and acceptable answers for each formulation.

\begin{figure}[h]
   \centering
   \captionsetup{width=4.5in}
   \begin{center}
      \includegraphics[height=2in]{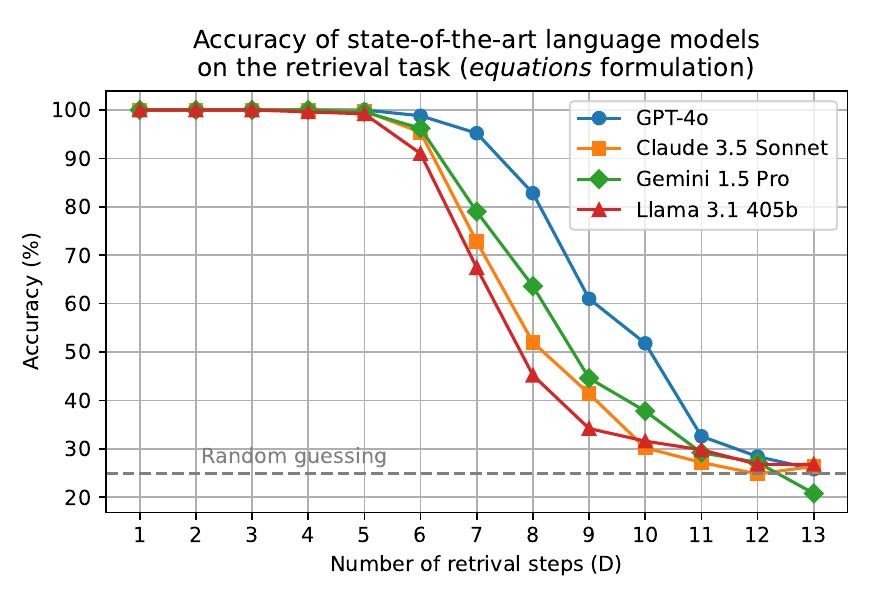}
      \includegraphics[height=2in]{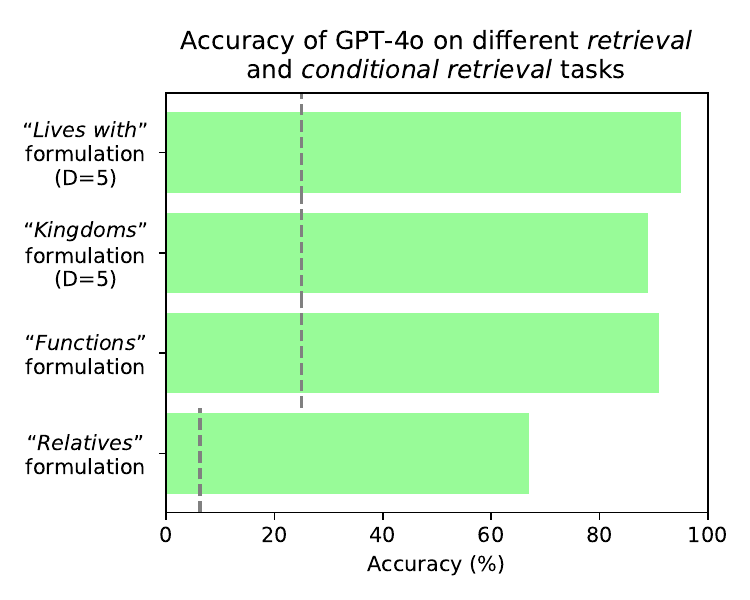}
   \end{center}
   \caption{Accuracy of large language models on the retrieval and
      conditional retrieval problems. Dashed lines indicate the accuracy of
      random guessing. Full prompts and benchmarking details are provided in
      Appendix \ref{prompts}.}
   \label{fig:llms}
\end{figure}

I test the large language models on 500 randomly generated questions for each
formulation. The results are presented in Figure~\ref{fig:llms}. For the
\textit{equations} formulation, I also measure the accuracy for different
difficulty levels $D$ (number of equations) and I find that large language
models can solve it almost perfectly for $D \leq 5$. Great performance is also
achieved on the \textit{lives-with} and \textit{kingdoms} formulations with
$D=5$, as well as on the conditional retrieval formulations \textit{functions}
and \textit{relatives}.

\section{Theoretical Analysis of Information Flow}
\label{sec:theory}

In this section, I theoretically establish that solving the retrieval problem
requires a minimum number of transformer layers that grows logarithmically with
the number of retrieval steps $D$. I model the information flow between
different positions during self-attention under the following simplifying
assumptions:

\textbf{Assumption 1.} During self-attention, a position can only attend to
another position if they already share a piece of information.

This assumption is motivated by the fact that a position can only attend to
another position if their key and query vectors align. Constructing aligned
key-query pairs is only possible if the two positions already share some
information. More precisely, the shared information must be located in the row
spaces of the query and key matrices of the attending and attended positions,
respectively.

\textbf{Assumption 2.}
When a position attends to another position, it retrieves all the information
contained in the attended position.

This assumption simplifies the analysis by ignoring network capacity
limitations. Solving this setting will give us a lower bound on the number of
layers required to solve the retrieval problem in the case of a limited network
capacity.

\begin{figure}[h]
   \begin{center}
      \includegraphics[width=4in]{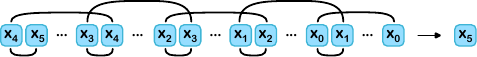}
   \end{center}
   \captionsetup{width=4in}
   \caption{Positions that contain shared information before any transformer
      layers in the case of $D=5$. Top edges denote shared token embeddings.
      Bottom edges denote shared positional encodings.}
   \label{fig:paths}
\end{figure}

Based on this simplified model of the attention mechanism, we can prove the
following result:

\textbf{Theorem 1.}
The last position in the sequence cannot retrieve the embedding vector $\vx_D$ of the target token
with $t$ transformer layers if $t < \log_3(2D)$.

This implies that at least $\log_3(2D)$ transformer layers are required to
solve a retrieval problem with $D$ retrieval steps. Note that this result is a
lower bound that does not take into account the limitations of network
capacity, causal masking, or training dynamics. In practice, we can expect the
number of required layers to be even higher.

I defer the complete proof to Appendix \ref{appendix:proof}. The intuition
behind this result comes from the fact that the last position in the sequence
cannot retrieve the target token $\vx_D$ without retrieving everything
in-between (i.e., $\vx_1$, $\vx_2$, $\ldots$, $\vx_{D-1}$). However, it is
possible to show that the number of retrieved tokens can grow at most by a
factor of 3 after each attention layer, hence the logarithmic growth of the
number of required layers.

\section{Minimal Problem Formulation}
\label{sec:minimal-formulation}

In order to better study the mechanism by which transformers solve the
retrieval problem, I introduce a minimal formulation of the retrieval problem
with $N$ retrieval chains, $D$ retrieval steps per chain, and $K$ embedding
dimensions. I use $N = 4$ and $K = 4$ throughout. Each chain contains $D+1$
unique symbols forming $D$ pairs and one query. For every input sequence, each
of the $N(D+1)$ unique symbols is assigned a $K$-dimensional vector whose
components are sampled i.i.d from a standard normal distribution.

I create the input sequences by perfectly interleaving the pairs of symbols
forming each retrieval chain, followed by the $N$ query symbols. I randomly
shuffle the query vectors. I also shuffle the input pairs from different chains
within the same retrieval step. Finally, I concatenate each token embedding
with a $K$-dimensional rotary positional encoding \citep{su2023rotary}. Each
input sequence will contain $N(2D+1)$ vectors of dimension $2K$. The output
sequences consist of $N$ vectors, one for each query token.

\section{Implicit Curriculum \& Number of Layers}
\label{sec:implicit-curriculum}

I consider two possible formulations: an implicit curriculum (IC) formulation
and a non-IC formulation. In the IC formulation, the target vectors have $DN$
dimensions and contain all the tokens forming each retrieval chain concatenated
(except the query token $\vx_0$). In the non-IC formulation, the target vectors
are $K$-dimensional and contain only the last token of each retrieval chain,
namely $\vx_D$.

\begin{figure}[h]
   \centering
   \captionsetup{width=4.6in}
   \begin{center}
      \includegraphics[width=2.7in]{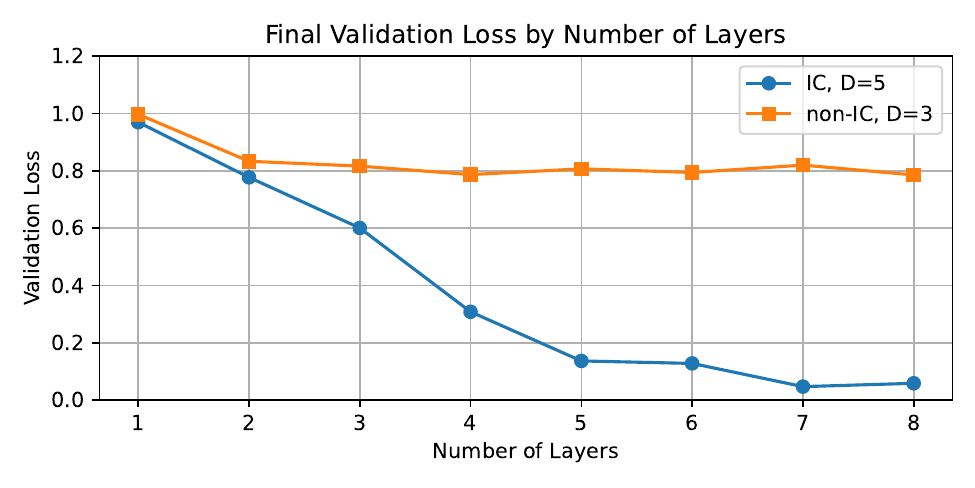}
      \includegraphics[width=2.7in]{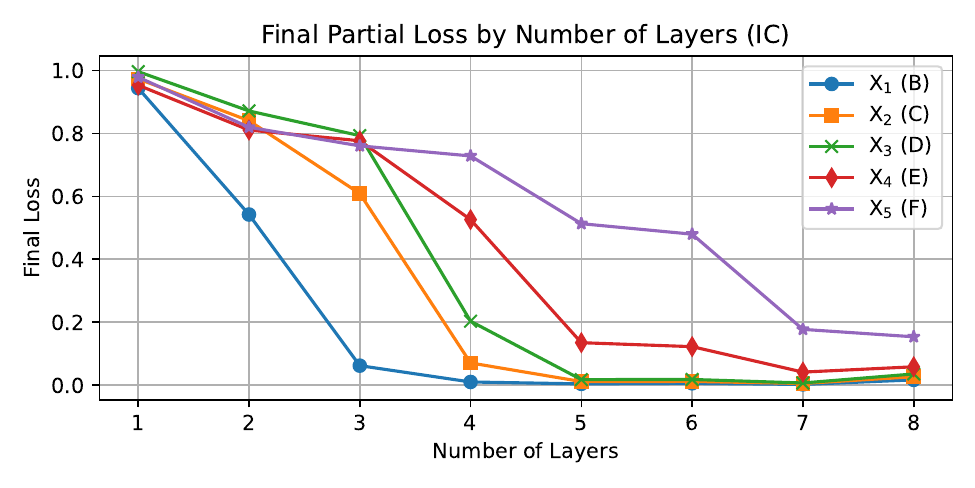}
   \end{center}
   \caption{Final validation loss by number of layers, averaged across multiple runs.
      Left: IC \textit{vs.} non-IC formulations. Right: Partial validation loss for each position in the retrieval chains (IC only).}
   \label{fig:mega_run}
\end{figure}

My initial experiments suggest that the implicit curriculum (IC) plays a
crucial role in the successful learning of the retrieval problem. To better
quantify this effect, I conduct two comprehensive sets of experiments, one for
each formulation (IC and non-IC). For each formulation, I train 64 transformers
with 1 to 8 layers (8 transformers for each number of layers). I plot the final
validation loss averaged across all runs with the same number of layers in
Figure~\ref{fig:mega_run} (left).

To better understand the connection between the number of layers and the
difficulty of the retrieval problem, I also plot the partial validation loss
for each position in the retrieval chains in the IC formulation
(Figure~\ref{fig:mega_run}, right). I use $D=5$ for the IC formulation to
better illustrate this connection, but only $D=3$ for the non-IC formulation to
illustrate the importance of the implicit curriculum.

\subsection{Training Details}

For each formulation and number of layers, I train 8 transformers following the
recipe of \citet{radford2019language}. Each transformer has 8 attention heads
per layer and residual streams of size 128. I train for 10k steps using the
Adam optimizer \citep{kingma2014adam} with a learning rate of $10^{-3}$,
decoupled weight decay of $0.1$ \citep{loshchilov2017decoupled}, a batch size
of 512, $2^{20}$ randomly generated training examples, layer normalization
\citep{ba2016layernormalization}, no dropout, and mean squared error loss. We
measure the final validation loss by averaging the validation loss over the
last 100 training steps.

\subsection{Results}

First, I observe that the IC formulation is essential for successful learning.
In the non-IC formulation, the transformers fail to learn the retrieval problem
even for $D=3$, regardless of the number of layers. I confirm that for 100\% of
the non-IC runs, the final validation loss is above $0.7$.

Second, I empirically confirm the connection between the number of layers and
the difficulty of the retrieval problem. For the IC formulation, the later
positions in the retrieval chains (corresponding to greater $D$) are more
difficult to learn and require more layers.

Third, I find our first hint regarding the emergence of retrieval heads. During
training with IC, the partial losses for earlier positions in the retrieval
chains always decrease faster than the partial losses for later positions. I
confirm that in 100\% of the IC runs, the partial loss goes below $0.5$ for
$\vx_1$ first, then for $\vx_2$, and so on. I will further investigate this
phenomenon in section \ref{head_emergence}.

\section{Reverse-Engineering the Circuits Learned}
\label{sec:reverse_engineering}

To understand the mechanism learned by transformers to solve the retrieval
problem, I train three transformers (denoted as A, B, and C) with 12 layers and
only one attention head per layer on the retrieval problem with $D=3$. I then
manually reverse-engineer the circuits learned by the transformers by studying
their attention maps. The uncovered circuits are depicted in
Figure~\ref{fig:circuits}. I describe my reverse-engineering process in detail
in Appendix \ref{appendix:reverse_engineering}.

\begin{figure}[h]
   \begin{center}
      \includegraphics[width=5.5in]{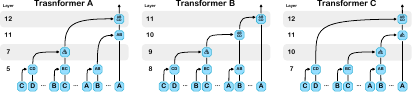}
   \end{center}
   \captionsetup{width=3.9in}
   \caption{Reverse-engineered circuits from three 12-layer transformers trained on the
      retrieval problem with $D=3$ and IC.}
   \label{fig:circuits}
\end{figure}

\subsection{Training Details}

I follow a similar training recipe as in the previous section. Each transformer
has 12 layers, one attention head per layer, and residual streams of size 128.
I train each transformer for 24k steps, a batch size of 128, and 262k randomly
generated training examples (IC).

\subsection{Results}

I find that transformers A, B, and C achieve a validation mean squared error of
less than $0.01$. By studying the attention maps in the trained transformers, I
observe that most attention heads do not perform any useful computation. Only a
few attention heads are responsible for the information flow. Their behavior is
easily interpretable (see Appendix \ref{appendix:attention_maps}).

I manually reverse-engineer the entire circuits learned by the three
transformers, which are depicted in Figure~\ref{fig:circuits}. I perform
extensive validations of the circuits using ablations. My reverse-engineering
process is described in detail in Appendix \ref{appendix:reverse_engineering}.

I observe two interesting facts about the reverse-engineered circuits:

\begin{enumerate}[label=\textbf{\roman*.}, itemsep=1pt]
   \item
         First, in all three transformers, the first relevant attention head is connecting the
         first and second tokens in each input pair, enabling the subsequent attention heads
         to attend to the second token in the pair using the value of the first token.
         This mechanism is highly reminiscent of the
         induction head mechanism \citep{olsson2022context, reddy2023mechanistic}.

   \item
         Second, except for the first attention head, the circuits learned by the transformers
         are very different. Interestingly, none of the transformers use the minimum
         number of attention heads required to solve the retrieval problem for $D=3$.
         All transformers use 4 attention heads, but it is possible to use only 3 (for
         example, by combining layers 7 and 11 in transformer A).
\end{enumerate}

\section{Emergence of Attention Heads during Training}
\label{head_emergence}

To better understand how the retrieval heads emerge during training, I train a
24-layer transformer (denoted as Transformer D) on the retrieval problem with
$D=4$ and IC. I manually reverse-engineer the circuits learned by Transformer D
following the same procedure described in Appendix
\ref{appendix:reverse_engineering}. Afterward, I measure the attention during
training for each attention path in the reverse-engineered circuit.

\begin{figure}[h]
   \centering
   \captionsetup{width=4.5in}
   \begin{center}
      \includegraphics[width=2.7in]{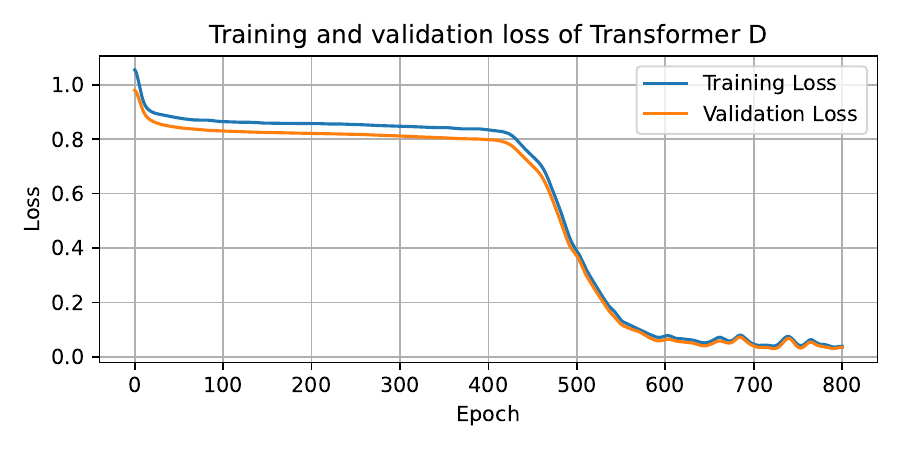}
      \includegraphics[width=2.7in]{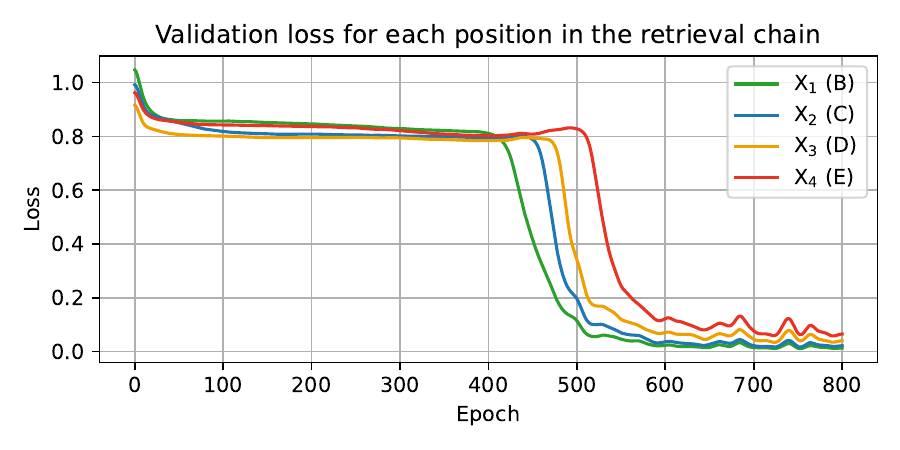}
   \end{center}
   \caption{Loss during training of Transformer D (24 layers) with IC and $D=4$.
      Left: training and validation loss. Right: partial validation loss for each position in the retrieval chain.}

   \label{fig:loss_D}
\end{figure}

\subsection{Training Details}

I follow a similar training recipe as in the previous sections. Transformer D
has 24 layers, one attention head per layer, and residual streams of size 512.
I train for 6400 steps (800 epochs), a batch size of 256, and 262k randomly
generated training examples (IC, $D=4$). To speed up the training and reduce
the checkpoint size, I remove the MLPs and reduce the head size to 16.

I save a checkpoint every 10 epochs (80 steps) that I later use to measure the
average attention for each attention path in the reverse-engineered circuit, at
each epoch during training, using 32 input sequences.

\begin{figure}[h]
   \begin{center}
      \includegraphics[width=5.1in]{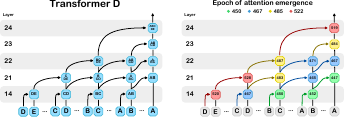}
   \end{center}
   \captionsetup{width=3.7in}
   \caption{Left: Reverse-engineered circuits of Transformer D.
      Right: the epoch when the average attention goes above $0.5$ for each attention path.}
   \label{fig:circuits_D}
\end{figure}

\subsection{Results}

Transformer D achieves a validation mean squared error of $0.031$. I plot the
training, validation, and partial validation loss in Figure~\ref{fig:loss_D}.
Using the same reverse-engineering procedure, I uncover a more complex circuit
than before, with multiple paths connecting the same positions
(Figure~\ref{fig:circuits_D}, left). After ablation, the mean squared error
increases to $0.045$.

For every checkpoint, I measure the average attention for each attention path
in the reverse-engineered circuit. I approximate the attention between
checkpoints using linear interpolation. I display the plots for each attention
path in Appendix \ref{appendix:attention_emergence}. Finally, I show the first
epoch when the average attention goes above $0.5$ for each attention path
(Figure~\ref{fig:circuits_D}, right).

By analyzing the partial loss curves and the emergence of attention paths, we
can make the following observations:

\begin{enumerate}[label=\textbf{\roman*.}, itemsep=1pt]
   \item
         After 450 epochs of slow learning, an induction head that
         can retrieve token \textbf{B} emerges
         abruptly on layers 14 and 21. This drives down the first
         partial loss.
   \item Quickly after, another attention head emerges on layer 22. This head reuses the
         induction head (with slight adjustments) to retrieve token \textbf{C} and drive
         down the second partial loss.
   \item Finally, two more heads emerge on layers 23 and 24 that reuse heads 14, 21, and
         22 to retrieve tokens \textbf{D} and \textbf{E}, respectively. This drives down
         the last two partial losses.
   \item Head 24 emerges much later than head 23, possibly due to the greater
         modifications required to reuse the existing circuit.
\end{enumerate}

Together, these observations strongly suggest the following possible
explanation for the importance of the implicit curriculum: \textit{The implicit
   curriculum provides a sequence of increasingly complex tasks that enables
   learning the entire retrieval mechanism one head at a time, starting with an
   induction head}.

\section{Discussion on Emergent Abilities}
\label{sec:discussion}

The retrieval problem has interesting connections to the emergent abilities of
large language models \citep{wei2022emergent}. An ability is \textit{emergent}
if it is not present in smaller models but is present in larger models.
Understanding emergence is an important direction because it could potentially
allow us to predict what abilities future models may have, as well as provide
new insights into how to train more capable language models.

The existence of tasks that require a minimum number of layers, such as the
retrieval problem, provides a possible explanation for the emergence of new
abilities in large language models. As the model grows in size, it becomes
possible to learn more complex circuits that would be impossible to learn in
smaller models. This unlocks new abilities that were previously unattainable.

\subsection{Emergent Abilities are Not a Mirage}

\citet{NEURIPS2023_adc98a26} have previously suggested that the emergence of
new abilities in large language models is just a ``mirage'' that appears only
under nonlinear or discontinuous metrics. My work provides a very strong
counterargument to this claim if we consider the ability of a model to solve
the retrieval problem. A transformer cannot solve the retrieval problem with
a specific difficulty unless it has the minimum number of necessary layers.

\subsection{The Implicit Curriculum of Natural Language}

Is it possible to understand how training on natural language data leads to
reasoning abilities? \citet{chan2022data} previously found that natural
language has specific data distributional properties that enable emergent
in-context learning in transformers. Is it possible to achieve a similarly
insightful understanding of the emergence of reasoning abilities in general,
not just in-context learning?

The retrieval problem provides a promising avenue for answering this question.
In Sections \ref{sec:implicit-curriculum} and \ref{head_emergence}, we saw that
retrieval heads can only emerge gradually, one by one, under the presence of an
implicit curriculum that provides a sequence of increasingly complex tasks.
This suggests that one property of natural language data is incredibly
important for the emergence of reasoning abilities: the presence of a very
diverse set of tasks with varying levels of difficulty.

\section{Related Work}

\paragraph*{Single-layer transformers.}

Perhaps the most well-studied setting for single-layer transformers is the
problem of modular addition. \citet{nanda2023progress} show that transformers
solve modular addition by arranging the embedding vectors in a circular
structure and leveraging the attention mechanism to perform trigonometric
operations. \citet{zhong2024clock} extend this work by uncovering other
algorithms and embedding structures. Even the training dynamics are beginning
to be understood, with \citet{ding2024survival} studying the survival of
initial circular representations and my previous work
\citep{musat2024clustering} proposing an effective theory of the training
dynamics by modeling the embeddings as a particle system.
\citet{quirke2023understanding} train a single-layer transformer with three
attention heads on the problem of n-digit integer addition. They find that
transformers break down the multi-digit addition task into parallel,
digit-specific streams, using different algorithms for various digit positions.

\paragraph*{Two-layer transformers.}

By studying two-layer transformers, \citet{olsson2022context} uncover a
mechanism termed \textit{induction head} that, given an input sequence
\textit{ab{\ldots}a}, can predict \textit{b}. One possible use of an induction
head is \textit{sequence copying}, but the authors argue that it can also
perform more high-level functions such as \textit{translation}. The induction
head is formed by two stacked attention heads. The first head copies into the
residual stream of \textit{b} the value of the previous token \textit{a}. The
second head is then able to attend to the token \textit{b} and copy it into the
residual stream of the final token. \citet{reddy2023mechanistic} explains the
emergence of the induction head during training by the sequential learning of
three nested logits enabled by an implicit curriculum.

\paragraph*{Large language models.}

Several studies on large language models use automated or semi-automated
methods to isolate circuits that solve a specific task
\citep{conmy2023towards,goldowsky2023localizing}. Such circuits often span many
layers, but their mechanisms remain not fully understood
\citep{wang2022interpretability}. Attention heads in large language models are
often strongly interdependent, which makes it difficult to isolate and
understand individual heads \citep{bricken2023monosemanticity}. In large
language models, even the simple task of greater-than comparison, which could
in principle be solved by a single-layer transformer, is solved by a complex
mechanism formed by multiple attention heads and MLPs \citep{hanna2024does}.

\section{Conclusion}

In this work, I introduced and studied the retrieval problem, a simple task
that requires transformers to retrieve information from multiple positions in
the input sequence. I showed that the retrieval problem requires a certain
number of layers to be solved. By training transformers on a minimal
formulation of the task, I found that transformers solve the task using a
mechanism that resembles an induction head. I found that this mechanism emerges
gradually with the help of an implicit curriculum, starting with an induction
head and then adding more heads one by one.

\paragraph*{Limitations.} My analysis of transformers trained on a minimal
formulation of the retrieval problem might not generalize perfectly to large
language models. I also do not provide a full explanation of the training
dynamics of transformers on the retrieval problem. Further research is needed
to fully understand the multi-layered circuits learned by large language models
and the training dynamics that enable their learning.

\paragraph*{Acknowledgments.} I would like to thank the anonymous reviewers for
their valuable feedback.

\newpage

\bibliography{iclr2025_conference}
\bibliographystyle{iclr2025_conference}

\newpage
\appendix

\section{Retrieval and Conditional Retrieval Prompts}
\label{prompts}

Below I provide one complete example for each of the retrieval and conditional
retrieval formulations used in the paper. The examples are generated using the
same programs used for benchmarking the large language models. Each example
consists of a prompt, a correct answer, and acceptable answers. Acceptable
answers are used to filter out incoherent answers by repeatedly sampling from
the model until an acceptable answer is found.

\subsection{Equations formulation (D = 5)}
\begin{lstlisting}[breaklines]
b = 2
c = 3
d = 0
a = 1
e = b
g = a
h = d
f = c
k = e
i = f
l = g
j = h
n = k
p = l
o = i
m = j
q = n
s = p
r = o
t = m
What is the value of s? Say directly only the numeric value, without any other words.

Correct: 1
Acceptable: 0, 1, 2, 3
\end{lstlisting}

\subsection{Lives-with formulation (D = 5)}
\begin{lstlisting}[breaklines]
Charlie lives in Cairo
David lives in Delhi
Alice lives in Berlin
Bob lives in Amsterdam
Henry lives with David
Eve lives with Charlie
Frank lives with Alice
Grace lives with Bob
Kate lives with Grace
Larry lives with Frank
Jack lives with Eve
Isabelle lives with Henry
Mary lives with Jack
Olivia lives with Isabelle
Nick lives with Kate
Peter lives with Larry
Rose lives with Peter
Queen lives with Nick
Tom lives with Olivia
Sam lives with Mary
Where does Rose live? Say directly only the name of the city, without any other words.

Correct: Berlin
Acceptable: Amsterdam, Berlin, Cairo, Delhi
\end{lstlisting}

\subsection{Kingdoms formulation}
\begin{lstlisting}[breaklines]
Bob lives in Silvania.
Alice lives in Novaria.
Charlie lives in Aurora.
David lives in Florinia.
Silvanians believe in celestianism.
Novarians believe in harmonianism.
Aurorans believe in elysianism.
Florinians believe in luminism.
Luminists eat beef.
Elysianists eat pork.
Harmonianists eat lamb.
Celestianists eat chicken.
Beef contains Astralyte.
Chicken contains Nephryon.
Lamb contains Zephyrium.
Pork contains Virellium.
Zephyrium causes Chronogy.
Astralyte causes Aetherflux.
Virellium causes Somnosis.
Nephryon causes Synthemia.
Who has Chronogy? Say directly the name without other words.

Correct: Alice
Acceptable: Alice, Bob, Charlie, David
\end{lstlisting}

\subsection{Functions formulation (conditional retrieval)}
\begin{lstlisting}[breaklines]
a(0) = 3
a(1) = 2
a(2) = 0
a(3) = 1
b(0) = 1
b(1) = 3
b(2) = 2
b(3) = 0
c(0) = 1
c(1) = 0
c(2) = 3
c(3) = 2
d(0) = 1
d(1) = 0
d(2) = 2
d(3) = 3
e = b
f = a
g = c
h = d
i = 0
j = 2
k = 3
l = 1
What is the value of f(i)? Say directly only the numeric value, without any other words.

Correct: 3
Acceptable: 0, 1, 2, 3
\end{lstlisting}

\subsection{Relatives formulation (conditional retrieval)}
\begin{lstlisting}[breaklines]
Penny lives in Canada.
Lily lives in Brazil.
Isabelle lives in France.
Cathy lives in Kenya.
George lives in Italy.
Adam lives in Mexico.
Kevin lives in Peru.
Ed lives in Laos.
Hank lives in Germany.
Mike lives in Japan.
Jane lives in England.
Fred lives in Hungary.
Dana lives in Norway.
Olivia lives in Qatar.
Bob lives in Denmark.
Nancy lives in Argentina.
John's mother is Jane.
John's sister is Olivia.
John's father is Ed.
John's brother is Mike.
Chris's mother is Penny.
Chris's sister is Dana.
Chris's father is Adam.
Chris's brother is George.
Diana's mother is Nancy.
Diana's sister is Isabelle.
Diana's father is Hank.
Diana's brother is Bob.
Eve's mother is Lily.
Eve's sister is Cathy.
Eve's father is Fred.
Eve's brother is Kevin.
Doctors live with their brothers.
Lawyers live with their mothers.
Teachers live with their sisters.
Engineers live with their fathers.
John works as a doctor.
Chris works as an engineer.
Diana works as a teacher.
Eve works as a lawyer.
Where does Eve live? Say directly only the name, without any other words.

Correct: Brazil
Acceptable: Argentina, Brazil, Canada, Denmark, England, France, Germany, Hungary, Italy, Japan, Kenya, Laos, Mexico, Norway, Peru, Qatar
\end{lstlisting}

\newpage
\section{Proof of Theorem 1 (minimum number of layers)}
\label{appendix:proof}

In this section, I provide the complete formal proof for \textbf{Theorem 1}
stated in Section \ref{sec:theory}. Recall the two assumptions underlying our
simplified model of the attention mechanism:

\textbf{Assumption 1.} During self-attention, a position can only attend to
another position if they already share a piece of information.

\textbf{Assumption 2.}
When a position attends to another position, it retrieves all the information
contained in the attended position.

\begin{figure}[h]
   \begin{center}
      \includegraphics[width=4in]{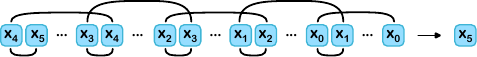}
   \end{center}
   \captionsetup{width=4in}
   \caption{Positions that contain shared information before any transformer
      layers in the case of $D=5$. Top edges denote shared token embeddings.
      Bottom edges denote shared positional encodings.}
   \label{fig:paths2}
\end{figure}

\textbf{Definition 1.} Let's consider all relevant input positions ordered by their reachability from
the last token following the paths of shared information before any attention layers, exactly as
depicted in Figure~\ref{fig:paths2} (e.g., $\vx_0 \vx_0 \vx_1 \vx_1 \vx_2 \vx_2 \ldots$ ).
Let's denote this sequence of positions as $\vs_i$ for $i \in \{0, 1, \ldots, 2D\}$.
In other words, $\vs_{2k}$ and $\vs_{2k+1}$ are the positions of the second and first occurrences of $\vx_k$, respectively.
For example, $\vs_0$ is the last position in the input sequence, $\vs_1$ is the first occurrence of $\vx_0$,
$\vs_2$ is the second occurrence of $\vx_1$, $\vs_{2D}$ is the position of the target token $\vx_D$, and so on.

\textbf{Definition 2.}
Let's denote as $\vr_{t,i}$ the residual stream at postition $\vs_i$ after layer $t$. Before any transformer layers, the residual stream contains only the token embedding and positional enconding:
\begin{equation}
   \vr_{0,i} = \vx_{\lfloor i/2 \rfloor} + \vp_{\lfloor (i + 1)/2 \rfloor} \quad \text{for all } i \in \{0, 1, \ldots, 2D\},
\end{equation}
where $\lfloor \cdot \rfloor$ denotes the floor function,
$\vx_k$ is the $k$-th token embedding, and $\vp_k$ is the positional encoding of the $k$-th input pair.

\textbf{Lemma 1.} During self-attention, a position can only attend to
another position if they already contain a shared token embedding or
positional encoding.

\textit{Proof.} In the retrieval problem, the token values and pair positions are
assigned randomly and independently. Knowledge of a token or position does not
provide any information about any other token or position. Therefore, the only
way for two positions to share information is if they already contain a shared
token embedding or positional encoding.

Note that before any attention layers, the position $\vs_i$ will contain
information shared only with positions $\vs_{i-1}$ and $\vs_{i+1}$. This is
because the token embedding $\vx_i$ is shared by $\vs_{2i}$ and $\vs_{2i+1}$,
while the positional encoding $\vp_i$ (for $i$-th input pair) is shared by
$\vs_{2i - 1}$ and $\vs_{2i}$.

\textbf{Definition 3.}
Let's denote as $\ve_i$ the piece of information shared by the positions $\vs_{i}$ and $\vs_{i+1}$
before any transformer layers. Specifically, we define $\ve_{i}$ for all $i \in \{0, \ldots, 2D\}$
such that $\ve_{2k} = \vx_k$ for $k \in \{0, 1, \ldots, D\}$
and $\ve_{2k-1} = \vp_k$ for $k \in \{1, \ldots, D\}$.

We are interested in the minimum number of layers $t$ such that $\vr_{t,0}$
might contain the target token $\vx_D$, also denoted as $\ve_{2D}$.

\textbf{Lemma 2.}
After every layer, every residual stream $\vr_{t,i}$ will contain a contiguous sequence of
pieces of information (e.g., $\{\ve_a, \ve_{a+1}, \ldots, \ve_b\}$).

\textit{Proof.}
We can show this using mathematical induction. The initial residual
stream $\vr_{0,i}$ contains only the token
embedding and the positional encoding, which represent the consecutive
pieces of information $\{\ve_{i}, \ve_{i + 1}\}$. During self-attention, the existing
contiguous sequence of pieces of information in $\vr_{t,i}$ will be merged with other
contiguous sequences (assumption 2) that share at least one piece of information
with $\vr_{t,i}$ (lemma 1). Their union in $\vr_{t+1,i}$ remains a contiguous sequence.

\textbf{Lemma 3.}
After every layer $t$, the contiguous sequence of pieces of information
in $\vr_{t,i}$ will have a length of at most $3^t + 1$ for all $i$.

\textit{Proof.}
We can show this using mathematical induction once again. The initial residual
streams $\vr_{0,i}$ contain exactly two pieces of information: the token embedding
and the positional encoding. During the $t$-th layer of self-attention, the
contiguous sequence of pieces of information in $\vr_{t-1,i}$ will grow by at most
$3^{t-1}$ pieces of information to the left and the right, resulting in a new total
length of at most $3^{t} + 1$.

\textbf{Theorem 1.}
The embedding vector $\vx_D$ of the target token cannot be present in the residual stream
$\vr_{t,0}$ after $t$ layers if $t < \log_3(2D)$.

\textit{Proof.}
The embedding of the target token $x_D$ corresponds to the piece of information
$\ve_{2D}$. For $\vr_{t,0}$ to contain $\ve_{2D}$, the length of its contiguous
sequence must be at least $2D + 1$. By Lemma 3, this length will not be reached
if $3^t + 1 < 2D + 1$, which is equivalent to $t < \log_3(2D)$.

\newpage
\section{Attention Maps}
\label{appendix:attention_maps}

\newboolean{showfigures}
\setboolean{showfigures}{true} % Change to show/hide figures (save compile time)

\foreach \tr in {A,B,C} {
      \subsection{Transformer \tr}

      \ifthenelse{\boolean{showfigures}}{
         \begin{center}
            \foreach \n in {0,...,11} {
                  \includegraphics[width=3cm]{attention_maps/model_\tr/layer_\n_head_0.pdf}
               }
         \end{center}
      }{}
   }
\newpage

\subsection{Transformer D}

\ifthenelse{\boolean{showfigures}}{
   \begin{center}
      \foreach \n in {0,...,23} {
            \includegraphics[width=3cm]{attention_maps/model_D/layer_\n_head_0.pdf}
         }
   \end{center}
}{}
\newpage

\section{Attention Emergence in Transformer D}
\label{appendix:attention_emergence}

\ifthenelse{\boolean{showfigures}}{
   \begin{center}
      \foreach \n in {1,...,13} {
            \includegraphics[width=4.5cm]{attention-emergence/plot_y\n.pdf}
         }
   \end{center}
}{}
\newpage

\section{Procedure for Reverse-Engineering Circuits}
\label{appendix:reverse_engineering}

As can be seen in Appendix \ref{appendix:attention_maps}, the attention maps
for each attention head in the transformers contain clear patterns that can be
be manually identified without the need for any additional tools. For this
reason, I decided to manually reverse-engineer the circuits, while validating
them thoroughly using ablations to ensure their correctness.

The exact procedure I follow to reverse-engineer the circuits:

\begin{enumerate}[label=\textbf{\arabic*.}, itemsep=1pt]
   \item I plot the attention maps for each transformer and each layer for different
         prompts.
   \item By observing the attention maps, I identify several possible mechanisms that
         could explain the attention patterns of each head.
   \item For each head, I determine which of the hypothesized mechanisms is correct
         using ablations (described below) and measuring the validation loss. I choose
         the simplest mechanism that maintains a low validation loss (below $0.05$)
         after ablation.
   \item I repeat steps 1-3 until the mechanism of each head has been identified.
   \item I validate the complete mechanism by performing combined ablations on all heads
         and measuring the validation loss.
   \item I validate that the uncovered mechanism is not excessive by attempting to
         further ablate all attention paths individually and measuring the validation
         loss.

\end{enumerate}

To validate the circuits, I measure the validation error after ablating the
attention maps in the following manner. For the attention heads that do not
perform any useful computation, I replace the attention weights with either
uniform attention or an identity matrix. For the attention heads that are
responsible for the information flow, I construct an attention map that is zero
everywhere except for the position that I expect the head to attend to, where
it is set to one.

After performing the combined ablations (step 5), I find that the mean squared
error increases slightly, but remains below $0.05$ for all transformers. By
further ablating any apparently useful attention path (step 6), the mean
squared error increases to $0.17-0.89$. The only exceptions are the first
useful layers of each transformer, which always attend to the previous
position. After ablating their attention as uniform, the error stays in the
range $0.05-0.1$, suggesting that they do not contribute directly to the final
output, but rather enable the information flow in the subsequent layers.

\newpage

\end{document}